\documentclass[a4paper]{article}

\usepackage{authblk}
\usepackage[utf8]{inputenc}
\usepackage[T1]{fontenc}
\usepackage{pslatex}
\usepackage[english]{babel}
\usepackage{fancyhdr}
\usepackage{hyperref}
\usepackage{amsmath,amssymb,amsfonts,amsthm}
\usepackage{graphicx}
\usepackage{amsmath}
\usepackage{amsthm}
\usepackage{amsfonts}
\usepackage{mathtools}
\usepackage{amssymb}
\usepackage{algorithm}
\usepackage{algorithmic}
\usepackage{microtype}
\usepackage{rotating}
\usepackage{todonotes}
\usepackage{tikz}
\usepackage{tikz-qtree}
\usepackage{xparse}
\usepackage{appendix}
\usepackage{url}
\usepackage{tabularx}
\usepackage{paralist}
\usepackage{booktabs}
\usepackage{multirow}
\usepackage{subcaption}
\usepackage{cancel}
\usetikzlibrary{calc,shapes.callouts,shapes.arrows}
\usetikzlibrary{automata}
\usetikzlibrary{positioning}
\usetikzlibrary{decorations.pathreplacing}
\usetikzlibrary{decorations.pathmorphing}
\usetikzlibrary{arrows}
\usetikzlibrary{shapes}

\newcommand{\F}{\mathbb{F}}

\newtheorem*{problem*}{Problem}

\DeclarePairedDelimiter\floor{\lfloor}{\rfloor}

\providecommand{\keywords}[1]{\textbf{\textit{Keywords }} #1}

\begin{document}

\title{A Systematic Evaluation of Evolving Highly Nonlinear Boolean Functions in Odd Sizes}

\author[1]{Claude Carlet}
\author[2]{Marko \DH urasevic}
\author[2]{Domagoj Jakobovic}
\author[3]{Stjepan Picek}
\author[4]{Luca Mariot}

\affil[1]{{\normalsize Department of Mathematics, Universit\'{e} Paris 8, 2 rue de la libert\'{e}, 93526 Saint-Denis Cedex, France}

    {\small \texttt{claude.carlet@gmail.com}}}

\affil[2]{{\normalsize Faculty of Electrical Engineering and Computing, University of Zagreb, Unska 3, Zagreb, Croatia} \\

{\small \texttt{marko.durasevic@fer.hr, domagoj.jakobovic@fer.hr}}}

\affil[3]{{\normalsize Digital Security Group, Radboud University, Postbus 9010, 6500 GL Nijmegen, The Netherlands} \\
	
	{\small \texttt{stjepan.picek@ru.nl}}}

\affil[4]{{\normalsize Semantics, Cybersecurity and Services Group, University of Twente, 7522 NB Enschede, The Netherlands} \\
	
	{\small \texttt{l.mariot@utwente.nl}}}
	
\maketitle

\begin{abstract}
Boolean functions are mathematical objects used in diverse applications. Different applications also have different requirements, making the research on Boolean functions very active. In the last 30 years, evolutionary algorithms have been shown to be a strong option for evolving Boolean functions in different sizes and with different properties. Still, most of those works consider similar settings and provide results that are mostly interesting from the evolutionary algorithm's perspective.

This work considers the problem of evolving highly nonlinear Boolean functions in odd sizes. While the problem formulation sounds simple, the problem is remarkably difficult, and the related work is extremely scarce.
We consider three solutions encodings and four Boolean function sizes and run a detailed experimental analysis. Our results show that the problem is challenging, and finding optimal solutions is impossible except for the smallest tested size. However, once we added local search to the evolutionary algorithm, we managed to find a Boolean function in nine inputs with nonlinearity 241, which, to our knowledge, had never been accomplished before with evolutionary algorithms.
\end{abstract}


\section{Introduction}
\label{sec:intro}

Boolean functions are used in diverse applications. For instance, some of the domains with long history and still active research developments include combinatorics~\cite{Rothaus}, coding theory~\cite{KERDOCK1972182}, computational complexity theory~\cite{10.5555/1540612}, and cryptography~\cite{Meaux2016}.
Since Boolean functions are used in various applications, the conditions they need to fulfill are different. For instance, depending on the application, it may be relevant to know the Boolean function size, whether it is even or odd, or the value for some specific property.
Thus, since there are multiple scenarios to consider, the research on Boolean functions is an active research domain. Considering how to construct Boolean functions with specific properties, two main directions are to either use algebraic constructions or heuristics~\cite{carlet_2021}.\footnote{It is also possible to use a hybrid approach where both algebraic constructions and heuristics are used, see, e.g.,~\cite{DBLP:conf/ppsn/PicekMBJ14}.} Algebraic constructions have the advantage of clear mathematical formulation, commonly working for multiple (all) sizes. On the other hand, heuristics need to be tested for every dimension of interest, but one could construct functions with properties that are not attainable with algebraic constructions (or, at least, with the currently known algebraic construction). Unfortunately, heuristics commonly struggle when considering larger Boolean functions since the search space has hyperexponential growth with respect to the function size: $2^{2^n}$.
This paper considers evolutionary algorithms (EA) to construct Boolean functions with high nonlinearity and odd sizes.

While this problem sounds simple, it is far from it. For instance, it was not known for more than 30 years if a Boolean function of nine variables could have a nonlinearity larger than 240. This problem was solved in 2007 with simulated annealing when Kavut et al. answered it positively~\cite{4167738}. However, the authors needed to use custom heuristics and limit the search space to a class of Boolean functions called rotation symmetric (RS) Boolean functions, which is considerably smaller than the search space size for general Boolean functions (see Table~\ref{tab:searchsizes}).

Considering the works done with evolutionary algorithms, we can informally divide them into those relevant from both mathematical and EA perspectives and those relevant from the EA side only. 
In the first category, we can single out the result from Kavut et al. since it solved a long-standing open problem~\cite{4167738}. Another example is the work from Carlet et al., where the authors used evolutionary algorithms to improve the algebraic construction of Boolean functions~\cite{10.1145/3449639.3459362}. Finally, some works consider evolving Boolean functions with properties that do not achieve good values when Boolean functions are constructed with algebraic constructions~\cite{DBLP:conf/ppsn/PicekBJ14,Wang2019}.
On the other hand, multiple papers have relevance from the evolutionary perspective only (e.g., in the sense of benchmarking) since the results either do not reach anything new (negative results) or there is no analysis that would show if results are new (for example, finding Boolean functions with desired properties but not evaluating if those functions are new or achievable by known algebraic constructions). Naturally, there is nothing wrong with considering only the evolutionary perspective. Still, one needs to be careful not to consider problems that also hold little relevance from the evolutionary perspective anymore. For instance, while evolving bent Boolean functions was challenging up to a few years ago, more recent EA results show the problem rather readily solvable~\cite{evolvingconstruction,10.1145/3067695.3084220}. Since we also know multiple algebraic constructions that produce bent Boolean functions, the problem relevance becomes less clear from both the mathematical and EA sides. 
Thus, while we can conclude that evolving Boolean functions is still very interesting, one must take care to find relevant problem instances. We concentrate on one such problem relevant to mathematical and EA research.

In this work, we systematically evaluate the problem of evolving highly nonlinear Boolean functions in odd dimensions (and small size). We emphasize that we do not constrain functions to be balanced, but we do not consider even sizes, making it impossible to obtain bent Boolean functions.
The constraint on high nonlinearity and odd dimension makes the problem mathematically relevant, as we do not know algebraic constructions capable of reaching upper bounds for a number of Boolean function sizes. Since the problem is difficult (due to the large search space), we limit our attention to small function sizes, allowing us to reach more relevant conclusions. 
More precisely, we evaluate three solution encodings (bitstring, floating-point (for floating-point, we investigate six algorithms), and symbolic) and four Boolean function sizes.
The results show that the problem is difficult: we can find optimal results for certain sizes, but such solutions are rare. Already for nine inputs, none of the algorithms can reach the optimal value. 
Then, we add several local search variants to our algorithms, making the results somewhat better, especially for the solution representation performing suboptimally before. Interestingly, with one evolutionary algorithm and local search combination, we even find a Boolean function with nonlinearity 241 and size 9. As far as we know, this is the first time EAs have found such a function.


\section{Preliminaries on Boolean Functions}
\label{sec:preliminaries}

We denote positive integers with $n$ and $m$: $n, m \in \mathbb{N}^+$. Moreover, we denote the Galois (finite) field with two elements as $\mathbb{F}_{2}$ and the Galois field with $2^n$ elements by $\mathbb{F}_{2^n}$. An $(n, m)$-function is a mapping $F$ from $\mathbb{F}_{2}^{n}$ to $\mathbb{F}_{2}^{m}$. 
When $m=1$, the function $f$ is called a Boolean function (of size $n$). 
We endow the vector space $\mathbb{F}_2^n$ with the structure of that field, since for every $n$, there exists a field $\mathbb{F}_{2^n}$ of order $2^n$ that is an $n$-dimensional vector space. The usual inner product of $a$ and $b$ equals $a\cdot b = \bigoplus_{i=1}^{n} a_{i}b_{i}$ in $\mathbb F_{2}^n$.

\subsection{Representations}

One common way to uniquely represent a Boolean function $f$ on $\mathbb{F}_{2}^{n}$ is by using its truth table. The truth table of a Boolean function $f$ is the list of pairs of function inputs (in $ \mathbb F_2^n$) and function outputs, with the size of the value vector being $2^n$. 
The output vector is the binary vector composed of all $f(x), x \in \mathbb{F}_2^n$, with a certain order selected on $\mathbb{F}_2^n$. 
Usually, one uses a vector $(f(0),\ldots, f(1))$ that contains the function values of $f$, ordered lexicographically~\cite{carlet_2021}. 

The Walsh-Hadamard transform $W_{f}$ is another commonly used unique representation of a Boolean function $f$. The Walsh-Hadamard transform measures the correlation between $f(x)$ and the linear functions $a\cdot x$, see, e.g.,~\cite{carlet_2021}
\begin{equation}
W_{f} (a) = \sum\limits_{x \in \mathbb{F}_{2}^{n}} (-1)^{f(x) + a\cdot x},
\end{equation}
with the sum calculated in ${\mathbb Z}$.

\subsection{Properties and Bounds}
\label{sec:boolean_properties}

\paragraph{Balancedness}
A Boolean function $f$ is balanced if it takes the value one exactly the same number of times ($2^{n-1}$) as the value zero when the input ranges over ${\mathbb F}_2^n$.

\paragraph{Nonlinearity}
The minimum Hamming distance between a Boolean function $f$ and all affine functions is the nonlinearity of $f$.
The nonlinearity $nl_{f}$ of a Boolean function $f$ can be calculated from the Walsh-Hadamard values~\cite{carlet_2021}:
\begin{equation}
\label{eq:nonlinearity}
nl_{f} = 2^{n - 1} - \frac{1}{2}\max_{a \in \mathbb{F}_{2}^{n}} |W_{f}(a)|.
\end{equation}

The Parseval relation $\sum\limits_{a\in {\mathbb F}_2^n}W_f(a)^2=2^{2n}$ implies that the nonlinearity of any $n$-variable Boolean function is bounded above by the so-called covering radius bound:
\begin{equation}
\label{eq_boolean_covering}
    nl_{f} \leq 2^{n-1}-2^{\frac n 2 - 1}.
\end{equation}
Eq.~\eqref{eq_boolean_covering} cannot be tight when $n$ is odd. 
The functions whose nonlinearity equals the maximal value from Eq.~\eqref{eq_boolean_covering} are referred to as bent, and they exist only for $n$ even.

For $n$ odd, a slightly better bound is $2\lfloor 2^{n-2}-2^{\frac n 2 - 2}\rfloor$~\cite{568715}.
The nonlinearity $2^{n-1} - 2^{\frac{n-1}{2}}$ is called the quadratic bound\footnote{Note that when we speak of a quadratic bound concerning general Boolean functions, it is not strictly speaking a bound but rather a value that we can try to exceed with the nonlinearity of certain functions.} since for $n$ odd, it is a tight upper bound on the nonlinearity of Boolean functions with algebraic degree at most two. 
Note also that it is called a bent concatenation bound since it is a tight upper bound on the nonlinearity of the concatenation of two bent functions $f, g$ in $n-1$ variables.
The quadratic bound is the best for $n\leq 7$ while for $n\geq 9$, better nonlinearity exists, see~\cite{carlet_2021}.
The detailed best-known (obtained) results are given in Table~\ref{tab:nl}.

\begin{table}
\scriptsize
  \centering
  \caption{Nonlinearities of Boolean functions in odd dimensions. The best-known results are taken from~\cite{carlet_2021}.}
  \label{tab:nl}
  \begin{tabular}{ccccc}
    &             \\
    \multicolumn{4}{c}{$n$}\\\toprule
    condition &  $7$  & $9$  & $11$  & $13$ \\ \midrule
    quadratic bound & 56 & 240 & 992 & 4032 \\\midrule
    best-known & 56 & 242 & 996 & 4040 \\\midrule
    upper bound & 58 & 244 & 1000 & 4050 \\\bottomrule
  \end{tabular}
\end{table}

\subsection{Rotation Symmetric Functions}

A Boolean function over $\mathbb{F}_2^n$ is called rotation symmetric (RS) if it is invariant under any cyclic shift of input coordinates: 
$$(x_0, x_1, \ldots,  x_{n-1}) \rightarrow (x_{n-1},  x_0, x_1, \ldots, x_{n-2}).$$

Clearly, the number of rotation symmetric Boolean functions will be less than the number of Boolean functions, as the output value remains the same for certain input values. 
Stănică and Maitra used the Burnside lemma to deduce that the number of rotation symmetric Boolean functions equals $2^{g_n}$, where $g_n$ equals~\cite{STANICA20081567}:
\begin{equation}
    g_n = \frac{1}{n}\sum_{t|n}\phi(t)2^{\frac{n}{t}},
\end{equation}
and $\phi$ is the Euler phi function (counting the number of positive integers less than $n$ that are coprime to $n$).
Thus, $g_n$ represents the number of orbits where an orbit is a rotation symmetric partition composed of vectors equivalent under rotational shifts.
 
We provide the number of orbits for the rotation symmetric Boolean functions in Table~\ref{tab:rots}. Clearly, already for $n=9$, exhaustive search is not an option.
\begin{table}
\scriptsize
  \centering
  \caption{The number of orbits for the rotation symmetric Boolean functions.}
  \label{tab:rots}
  \begin{tabular}{ccccc}
    &             \\
    \multicolumn{4}{c}{$n$}\\\toprule
    variables&	7&	9&	11&	13 \\\midrule
    $g_n$&	20&	60&	188&	632 \\
\bottomrule
  \end{tabular}
\end{table}
Clearly, considering rotation symmetric functions allows an exhaustive search for larger Boolean function sizes or at least a ``simpler'' problem for heuristics. We list the search space sizes in Table~\ref{tab:searchsizes}.

\begin{table}
\scriptsize
  \centering
  \caption{The number of (rotation symmetric) Boolean functions.}
  \label{tab:searchsizes}
  \begin{tabular}{ccccc}
    &             \\
    \multicolumn{4}{c}{$n$}\\\toprule
    criterion & $7$  & $9$  & $11$ & $13$ \\ \midrule
    \# general  & $2^{128}$ & $2^{512}$  & $2^{2048}$  & $2^{8192}$ \\\midrule
    \# RS & $2^{20}$  & $2^{60}$  & $2^{188}$  & $2^{632}$ \\
\bottomrule
  \end{tabular}
\end{table}

More information about Boolean functions can be found in, e.g.,~\cite{MacWilliams-Sloane,carlet_2021}.

\section{Related Work}
\label{sec:related}

As already discussed, most of the research works on evolutionary algorithms and Boolean functions considers two cases: 1) evolving bent Boolean functions (thus, considering only even dimensions and imbalanced Boolean functions) or 2) evolving balanced, highly nonlinear Boolean functions (plus maybe some other cryptographic properties)~\cite{Djurasevic2023}.
On the other hand, evolving maximally nonlinear Boolean functions in odd dimensions is a relatively unexplored topic.
We provide a brief overview of works considering highly nonlinear Boolean functions in odd sizes, novel EA techniques, or constructing rotation symmetric Boolean functions.

To our knowledge, Millan et al. were the first to apply a genetic algorithm (GA) to evolve Boolean functions with good cryptographic properties~\cite{10.1007/BFb0028471}. 
The authors used a genetic algorithm (and hill climbing) to evolve Boolean functions with high nonlinearity. The authors considered sizes from 8 to 16 inputs, and the best result for nine inputs is achieved with a combination of GA and hill climbing and it equals 236.
Clark and Jacob used a combination of simulated annealing and hill-climbing with a cost function motivated by the Parseval theorem to find functions with high nonlinearity and low autocorrelation~\cite{10.1007/10718964_20}. Most of the work considers functions with eight inputs, but the authors also report results for sizes 4 to 12. The best result for the nine inputs and genetic algorithms equals 236, and simulated annealing is 238.
Burnett et al. used custom heuristics to generate Boolean functions with good cryptographic properties~\cite{burnett}. They reported a nonlinearity of 240 for the Boolean function in nine inputs.

Picek et al. were the first to use genetic programming (GP) to find Boolean functions with high nonlinearity (alongside more properties)~\cite{10.1145/2464576.2464671}. The authors considered only Boolean functions with eight inputs, but already from there, it was clear that GP could easily outperform GA.
Mariot and Leporati proposed using Particle Swarm Optimization to find Boolean functions with good trade-offs of cryptographic properties~\cite{10.1145/2739482.2764674}. The authors consider sizes between 7 and 12 variables and report the best nonlinearity of 236 for the Boolean function with nine inputs.

Stănică et al. used simulated annealing to evolve rotation symmetric Boolean functions~\cite{StanicaMC04}. By reducing the search space this way, the authors constructed Boolean functions in nine variables with nonlinearity 240.
Kavut et al. used a steepest descent-like iterative algorithm to construct highly nonlinear Boolean functions~\cite{4167738}. The authors found imbalanced Boolean functions in nine variables with a nonlinearity of 241. This represented a breakthrough as, before this result, it was not known if one could obtain a function in nine variables with nonlinearity larger than 240.
Kavut et al. conducted an efficient exhaustive search of rotation symmetric Boolean functions in nine variables having nonlinearity greater than 240~\cite{Kavut2006}. They showed there are 1512 functions with a nonlinearity of 241 and no rotation symmetric Boolean function with a nonlinearity greater than 241.
Kavut and Yucel used a steepest-descent-like iterative algorithm to construct imbalanced Boolean functions in $9$ variables with nonlinearity $242$~\cite{KAVUT2010341}. For this result, the authors considered the generalized rotation symmetric Boolean functions class to allow nonlinearity to potentially reach above 241 while, at the same time, making the search space size significantly smaller than for general Boolean functions.
Liu and Youssef used simulated annealing in combination with some algebraic techniques to construct balanced rotation symmetric Boolean functions in 10 inputs with nonlinearity equal to 488~\cite{4729749}. 
Wang et al. employed genetic algorithms (GAs) to construct rotation symmetric Boolean functions~\cite{Wang2022}. The authors reported constructing balanced, highly nonlinear rotation symmetric functions in 8, 10, and 12 inputs.
Recently, Carlet et al. investigated evolutionary algorithms for the evolution of rotation symmetric Boolean functions~\cite{carlet2023new}. The authors report finding balanced Boolean functions in nine variables with nonlinearity 240. Moreover, they achieve it in two ways: either evolving general Boolean functions with the tree encoding or evolving rotation symmetric Boolean functions with the bitstring encoding. Unfortunately, they do not consider the evolution of imbalanced Boolean functions in odd sizes.

To conclude, a large part of the related works either completely ignore the Boolean functions in odd sizes or impose constraints that the functions need to be balanced. From the remaining works, most achieve suboptimal results (i.e., not reaching the upper bound). 
The best results are achieved with custom heuristics, and evolutionary algorithms do not seem to be able to compete.

\section{Experimental Settings}
\label{sec:settings}

\subsection{EC Representations}

\textbf{Bitstring Encoding.}
The most common option for encoding a Boolean function is the bitstring representation~\cite{Djurasevic2023}. The bit string represents the truth table of the function. For a Boolean function with $n$ inputs, the truth table is coded as a bit string with a length of $2^n$.
For rotationally symmetric Boolean functions, the number of truth table entries to be coded is significantly lower and is specified in Table~\ref{tab:rots}.
For each evaluation, the bitstring genotype is first decoded into the full Boolean truth table (which is trivial since we know the orbits), and the desired property is computed (since we consider the nonlinearity property, we must first translate the truth table representation into the Walsh-Hadamard spectrum).

\textbf{Symbolic Encoding.}
The second approach in our experiments uses tree-based GP to represent a Boolean function in its symbolic form.
In this case, we represent a candidate solution by a tree whose leaf nodes correspond to the input variables $x_1,\cdots, x_n \in \F_2$. The internal nodes are Boolean operators that combine the inputs received from their children and forward their output to the respective parent nodes.

The output of the root node is the output value of the Boolean function. The corresponding truth table of the function $f: \F_2^n \to \F_2$ is determined by evaluating the tree over all possible $2^n$ assignments of the input variables at the leaf nodes. 
Each GP individual is evaluated according to the truth table it generates.

\textbf{Floating Point Encoding.}
The last approach to representing a Boolean function is the floating-point genotype, which is defined as a vector with continuous variables.
This requires defining the translation of a vector of floating point numbers into the corresponding genotype, which is then translated into a complete truth table (binary values).
The idea behind this translation is that each continuous variable (a real number) of the floating point genotype represents a subsequence of bits in the genotype.
All real values in the floating point vector are restricted to the interval $[0, 1]$.
If the genotype size is $g_n$, the number of bits represented by a single continuous variable of the floating point vector can vary:
\begin{equation}
\label{eq:decode}
decode = \frac{g_n}{dimension},
\end{equation}
where the parameter $dimension$ denotes the floating point vector size (number of real values). This parameter can be modified if the genotype size is divisible by its value. 
The first step of the translation is to convert each floating point number to an integer value.
As each real value must represent $decode$ bits, the size of the interval decoding to the same integer value is given as:
\begin{equation}
interval = \frac{1}{decode}.
\end{equation}
To obtain a distinct integer value for a given real number, every element $d_i$ of the floating point vector is divided by the calculated interval size, generating a sequence of integer values:
\begin{equation}
int\_value_i = \floor*{\frac{d_i}{interval}}.
\end{equation}
The final translation step involves decoding the integer values into a binary string that can be used for evaluation. For further details on using floating point representation for evolving rotation symmetric Boolean functions, see~\cite{carlet2023new}.

\subsection{Fitness Function}
\label{subsec:fit}

Several objective functions can be defined to optimize Boolean function nonlinearity regardless of the representation and search algorithm. The fitness function used here was selected based on the literature study of common choices in related works~\cite{Djurasevic2023} and the authors' previous experience.
Apart from maximizing the nonlinearity value, the applied fitness function considers the whole Walsh-Hadamard spectrum and not only its extreme value (see Eq.~\eqref{eq:nonlinearity}).
Here, we count the number of occurrences of the maximal absolute value in the spectrum, denoted as $\#max\_values$.
As higher nonlinearity corresponds to a lower maximal absolute value, we aim for as few occurrences of the maximal value as possible, hoping it would be easier for the algorithm to reach the next nonlinearity value.
In this way, we provide the algorithm with additional information, making the objective space more gradual.
With this in mind, the fitness function is defined as:
\begin{equation}
\label{eq:second}
fitness = nl_{f} + \frac{2^n - \#max\_values}{2^n}.
\end{equation}
The second term never reaches the value of $1$ since, in that case, we effectively reach the next nonlinearity level.

\subsection{Algorithm Parameters}

\textbf{Bitstring Encoding.}
The corresponding variation operators we use are the simple bit mutation and the shuffle mutation.
The simple bit mutation inverts a randomly selected bit. The shuffle mutation shuffles the bits within a randomly selected substring.
For the crossover operators, we use the one-point crossover and uniform crossover. 
The one-point crossover combines a new solution from the first part of one parent and the second part of the other parent with a randomly selected breakpoint.
The uniform crossover randomly selects one bit from both parents at each position in the child bitstring that is copied.
Each time the evolutionary algorithm invokes a crossover or mutation operation, one of the previously described operators is randomly selected.

\textbf{Symbolic Encoding.}
In our experiments, we use the following function set: OR, XOR, AND, AND2, XNOR, IF, and function NOT that takes a single argument.
The function AND2 behaves the same as the function AND but with the second input inverted. 
The function IF takes three arguments and returns the second one if the first one evaluates to true and the third one otherwise. This function set is common when dealing with the evolution of Boolean functions with cryptographic properties~\cite{Djurasevic2023,carlet2023new}.

The genetic operators used in our experiments with tree-based GP are simple tree crossover, uniform crossover, size fair, one-point, and context preserving crossover~\cite{poli08:fieldguide} (selected at random), and subtree mutation.
The option to use multiple genetic operators was based on the initial experiments.

We employ the same evolutionary algorithm for both bitstring and symbolic encoding: a steady-state selection with a 3-tournament elimination operator (denoted SST). 
In each iteration of the algorithm, three individuals are chosen at random from the population for the tournament, and the worst one in terms of fitness value is eliminated. 
The two remaining individuals in the tournament are used with the crossover operator to generate a new child individual, which then undergoes mutation with individual mutation probability $p_{mut} = 0.5$. Finally, the mutated child replaces the eliminated individual in the population.

\textbf{Floating Point Encoding.}
When FP encoding is used, one can vary the number of bits a single FP value will represent ($decode$, Eq.,~\eqref{eq:decode}). 
Based on related work~\cite{carlet2023new}, all FP-based algorithms use the same setting with $decode = 3$.
The floating point representation can be used with any continuous optimization algorithm, which increases its versatility. 
In our experiments, we investigated the following algorithms: Artificial Bee Colony (ABC)~\cite{karaboga2014comprehensive}, Clonal Selection Algorithm (CLONALG)~\cite{brownlee2007clonal}, CMA-ES~\cite{hansen2003reducing}, Differential Evolution (DE)~\cite{pant2020differential}, Optimization Immune Algorithm (OPTIA)~\cite{cutello2006real}, and a GA-based algorithm with steady-state selection (GA-SST), which is also used with TT and GP and whose behavior is described above.
Due to lack of space, we do not provide algorithms' parameters but note we used the ECF software framework\footnote{Evolutionary Computation Framework,~\url{http://solve.fer.hr/ECF/}.} with default parameter values.

\section{Experimental Evaluation}
\label{sec:results}

\subsection{Evolutionary Algorithms}

\begin{figure*}[!ht]
     \centering
     \begin{subfigure}[b]{0.4\textwidth}
         \centering
         \includegraphics[width=\textwidth]{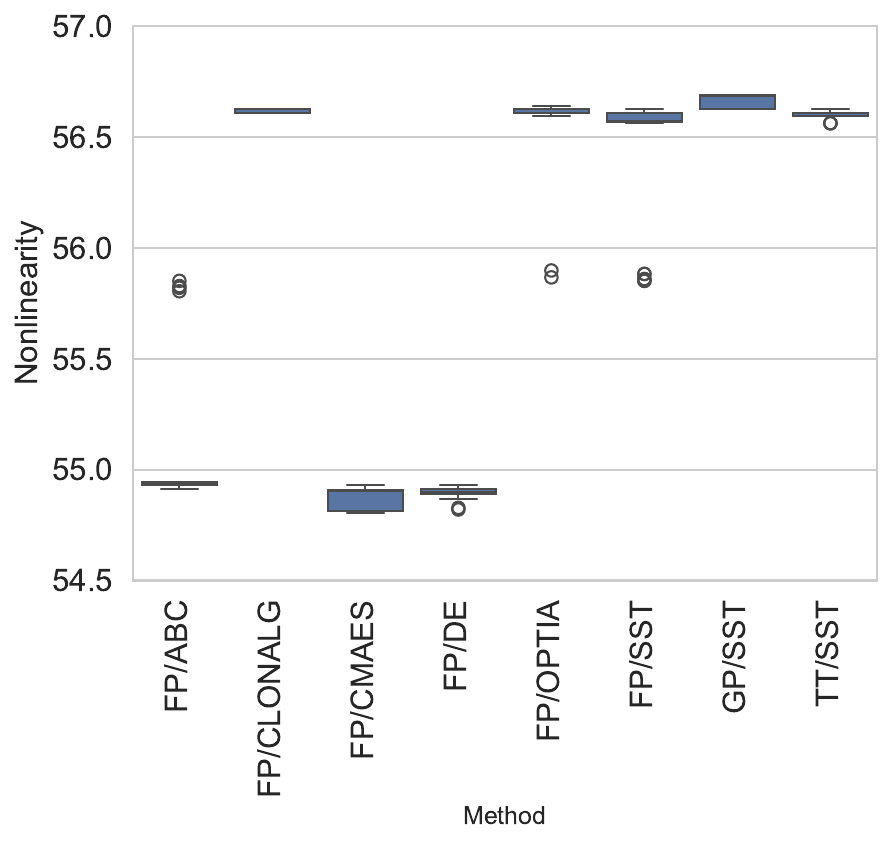}
         \caption{Size 7}
         \label{fig:box7}
     \end{subfigure}
     \begin{subfigure}[b]{0.4\textwidth}
         \centering
         \includegraphics[width=\textwidth]{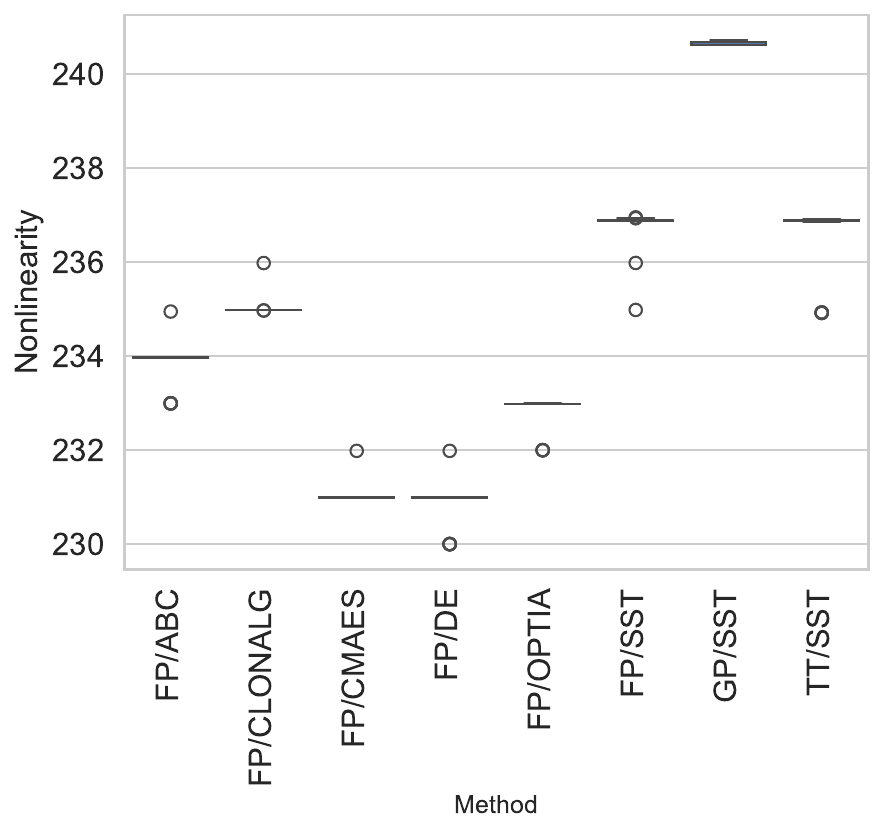}
         \caption{Size 9}
         \label{fig:box9}
     \end{subfigure}
     \begin{subfigure}[b]{0.4\textwidth}
         \centering
         \includegraphics[width=\textwidth]{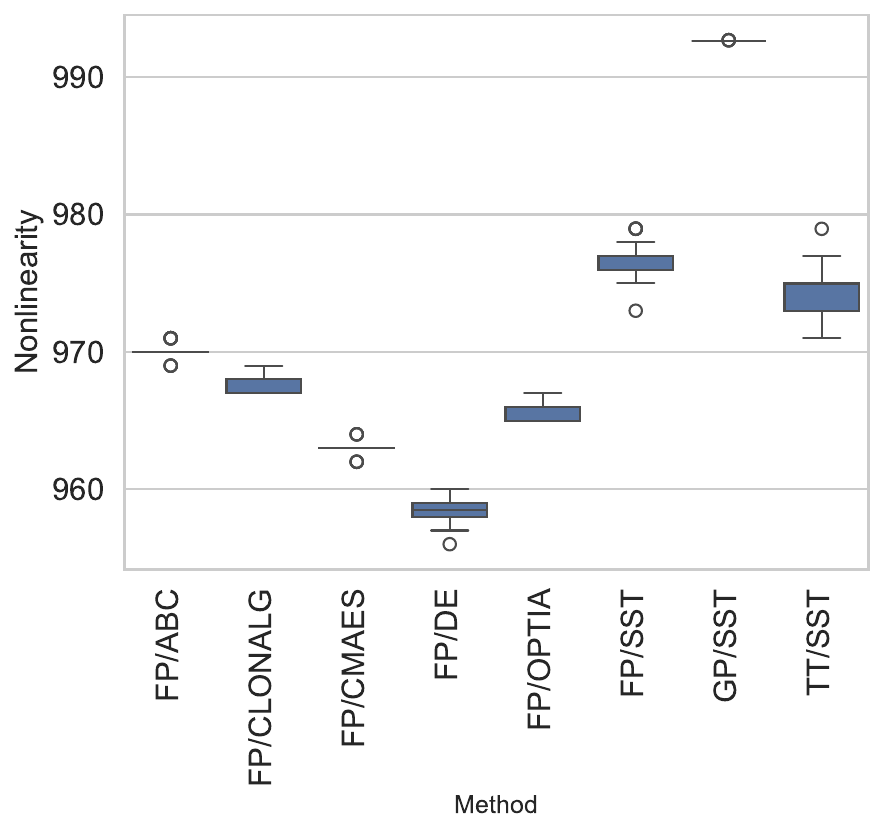}
         \caption{Size 11}
         \label{fig:box11}
     \end{subfigure}
     \begin{subfigure}[b]{0.4\textwidth}
         \centering
         \includegraphics[width=\textwidth]{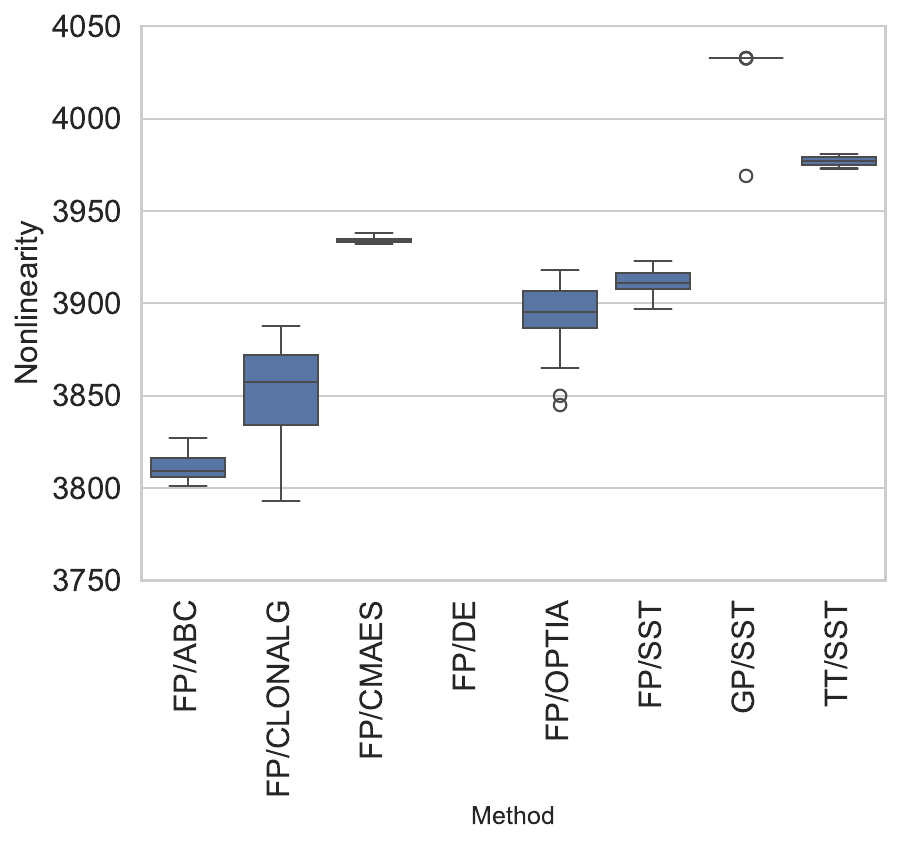}
         \caption{Size 13}
         \label{fig:box13}
     \end{subfigure}
       \caption{Boxplot representation of the results obtained for various sizes.}
        \label{fig:three graphs}
\end{figure*}

The results for the three representations and the different optimization methods are outlined in Table~\ref{tbl:res} for the four selected problem sizes. 
Each experiment was executed 30 times, with descriptive statistics like the maximum, average, and standard deviation being outlined in the table. 
To better outline the best results, they are given in bold for each problem size.
Furthermore, to better showcase the differences between the experiments, the results of all methods are also outlined in Figure~\ref{fig:three graphs} as boxplots.
In the case of size 13, the figure does not include the results for FP/DE due to the poor solutions the methods obtain, making it unreadable. 

The table and figure evidently show that the best results are obtained by GP across all the problem sizes. 
The differences between it and the other methods become more pronounced as the size of the problem increases. 
Regarding the TT and FP representations, it is not possible to say whether each one is consistently better than the other since it depends on the problem size. 
Regarding the optimization algorithms that were used with the FP representation, we see that the choice of the algorithm had a significant influence on the obtained results. 
Again, no single optimization method consistently achieved the best results for all problem sizes. However, the SST algorithm seems to be the most stable because it achieved good performance for most problem sizes. 
An additional benefit of GP against the other methods is that it achieved a very small standard deviation value, in several cases the smallest among all the methods. Thus, the GP results are not dispersed, and the algorithm is rather stable.
Regarding the comparison with the best-known solutions, for size 7, each representation obtained the best-known result, a solution with a nonlinearity of 56.
However, this was not the case for larger sizes, and the obtained solutions are worse than the best-known solutions for those sizes (see Table~\ref{tab:nl}). 

\begin{table*}
\tiny
\centering
\caption{Summary of the results of the various representations and optimization algorithms (obtained fitness values).}
\label{tbl:res}
\begin{tabular}{@{}llccc|ccc|ccc|ccc@{}}
\toprule
\multirow{2}{*}{Rep.} & \multirow{2}{*}{Alg.} & \multicolumn{3}{c}{7} & \multicolumn{3}{c}{9}  & \multicolumn{3}{c}{11} & \multicolumn{3}{c}{13}    \\ \cmidrule(l){3-14} 
                                &                            & max.   & avg.  & std. & max.   & avg.   & std. & max.   & avg.   & std. & max.    & avg.    & std.  \\ \midrule
\multirow{7}{*}{FP}             & ABC                        & 55.85  & 55.05 & 0.31 & 234.95 & 233.84 & 0.42 & 971.00 & 970.09 & 0.61 & 3827.00 & 3810.97 & 6.71  \\
                                & CLONALG                    & 56.63  & 56.62 & 0.01 & 235.98 & 235.01 & 0.18 & 969.00 & 967.76 & 0.57 & 3888.00 & 3853.40 & 24.37 \\
                                & CMAES                      & 54.93  & 54.65 & 0.50 & 231.98 & 231.02 & 0.18 & 964.00 & 963.00 & 0.52 & 3938.00 & 3934.23 & 1.41  \\
                                & DE                         & 54.93  & 54.90 & 0.03 & 231.98 & 230.79 & 0.48 & 960.00 & 958.50 & 1.01 & 2836.00 & 2701.12 &  58.23     \\
                                & OPTIA                      & 56.64  & 56.57 & 0.19 & 232.99 & 232.85 & 0.34 & 967.00 & 965.43 & 0.57 & 3918.00 & 3894.17 & 18.22 \\
                                & SST                        & 56.63  & 56.46 & 0.30 & 236.95 & 236.80 & 0.38 & 978.97 & 976.78 & 1.42 & 3923.00 & 3911.70 & 7.20  \\ \midrule
GP                              & SST                        & 56.69  & \textbf{56.64} & 0.03 & 240.72 & \textbf{240.64} & 0.03 & 992.69 & \textbf{992.63} & 0.02 & 4032.69 & \textbf{4030.52} & 11.62 \\ \midrule
TT                              & SST                        & 56.63  & 56.60 & 0.02 & 236.91 & 236.55 & 0.74 & 978.96 & 974.44 & 1.88 & 3980.99 & 3977.22 & 2.51  \\ \bottomrule
\end{tabular}
\end{table*}

To determine whether a statistically significant difference between the results obtained by different methods exists, we used the Kruskal-Wallis test with a significance value of 0.05.
Since a p-value of 0 was obtained for all problem sizes, we conclude that there is a significant difference in the performance of the tested algorithms and representations.
The post-hoc Dunn test with the Bonferroni correction method was used to determine these differences.

For problem size 7, the results demonstrate that although GP achieves the best result, it does not perform significantly better than the FP representation with the CLONALG or OPTIA methods.  
On the other hand, the TT representation was significantly worse than GP but equally good as FP for some algorithms (again, CLONALG and OPTIA).
For problem sizes 9 and 11, GP achieves significantly better results than all other methods, except for FP with SST, which came second.
Furthermore, in those cases, there is no difference between the TT and FP representation (when considering the result obtained by the best algorithm).
For size 13, GP achieves significantly better results than all other methods, except for TT, against which the differences are not significant. 
Furthermore, TT again achieves equally good results as the FP representation with CMAES.

Based on the previously outlined observations and analyses, we can conclude that GP is the most appropriate method for tackling the considered problem among those tested. It consistently achieved the best results, which were significantly better than most of the results obtained by any other method. 
Unfortunately, even these best results mostly fall short of the result with, e.g., custom heuristics.

\subsection{Evolutionary Algorithms + Local Search}

Since the case with nine variables was the smallest size where evolutionary methods did not obtain the best-known value, we investigated the possibilities of improving the efficiency in this particular case.
The first modification adds local search, which was applied in two forms.
The first form is a mutation-based local search operator: the operator acts on a single solution and performs a number of mutations. 
If a better solution is found, the new solution immediately replaces the original one, and the operator is applied again.
If no better solution is found after a predefined number of mutations, the operator terminates.
The operator is applied after each generation and acts upon the current best solution and a number of random solutions.
In our experiments, the number of solutions undergoing local search was set to 5\% of the population size, and the number of trials (random mutations per individual) was set to 25.
This operator is general in that it can be applied to any encoding.

However, for the bitstring representation, we included the second form of the local operator that performs individual bit flips instead of random mutations.
The operator is exhaustive, meaning it will perform all possible bit flips and terminate only if there is no improvement.

We applied the local search operators only to GP and bitstring encodings as the most efficient variants; in the bitstring case, three combinations were tested with either the mutation (denoted as "-LS1") or bit flip operator (denoted as "-LS2"), or both (denoted as "-LS3").

The results with these modifications were not encouraging, despite the experiment design in which we executed a thousand runs for every combination with a time limit of 2000 seconds, which amounts to approximately 300 million evaluations per run.
The GP efficiency was not altered since GP always found the same nonlinearity value of 240 in every run, with or without local search. The results for the bitstring encoding were slightly improved and are shown in Table~\ref{tbl:ls}.


Following these experiments, we introduced another modification: the use of rotation invariant encoding.
In this encoding, we only consider rotation-symmetric Boolean functions, limiting the number of possible solutions with high nonlinearity while drastically reducing the search space.
For instance, in the nine variable case, the representation for rotation symmetric functions consists of only 60 bits (as opposed to 512 in the general case).
These results are also included in Table~\ref{tbl:ls} and denoted with "-RI".

To better outline the effect of different LS operators on the results, Figure~\ref{tbl:ls} provides the boxplot representation of the results.
Clearly, applying LS operators affects the results, especially the application of the one based on the bit flip operator.
To determine whether the improvement in the results is statistically significant, the Kruskal-Wallis test was again applied. 
Since a p-value of 0 was obtained, we deduce that the results are significantly different.
The post hoc analysis demonstrates that by using LS, it is possible to improve the results significantly compared to the basic algorithm. 
Furthermore, the rotation invariant algorithm variant leads to significant improvements in the results. Remarkably, it also archives the nonlinearity of 241 for the TT-RI-LS1 combination. To our knowledge, this is the first time EA reached nonlinearity 241 for Boolean functions with nine inputs. 
Regarding the different LS operators, the analysis demonstrates that there is no significant difference between the three operator types. 

Let us consider how our results compare with~\cite{4167738}. The authors reported that among 200 million RSBFs in nine inputs evaluated with the steepest descent-like algorithm, five have nonlinearity 241. Clearly, this is more successful than what we achieved, as we found only one such function. The question is from where the better performance arises. First, there is a difference in the objective function, but they are rather similar since Kavut et al. considered the sum of square errors of the Walsh-Hadamard values. Next, the difference is in the steepest descent-like nature of their algorithm since they introduced a step where, once the cost cannot be minimized further, there is a deterministic step in the reverse direction corresponding to the smallest possible cost increase. Finally, we did not use exclusively local search but a combination with EA. Based on the results, we conclude that local search is a crucial step, which indicates that EA operators are either 1) too disruptive or 2) reach local optima and cannot produce a small change required to improve the fitness value (increasing nonlinearity from 240 to 241 requires (in the best case scenario), only a single change in the truth table representation). 

\begin{table}
\scriptsize
\centering
\caption{Results for the TT representation with LS operators}
\label{tbl:ls}
\begin{tabular}{@{}lccc@{}}
\toprule
           & max.   & avg.   & std. \\ \midrule
TT         & 236.91 & 236.55 & 0.74 \\
TT-LS1    & 238.83 & 237.98 & 0.95 \\
TT-LS2    & 238.87 & 238.58 & 0.64 \\
TT-LS3    & 238.87 & 238.69 & 0.48 \\ \midrule
TT-RI-LS1 & \textbf{241.75} & 240.75 & 0.05 \\
TT-RI-LS2 & 240.88 & 240.80 & 0.03 \\
TT-RI-LS3 & 240.90 & 240.79 & 0.04 \\ \bottomrule
\end{tabular}
\end{table}

\begin{figure}[!ht]
    \centering
    \includegraphics[width=0.8\columnwidth]{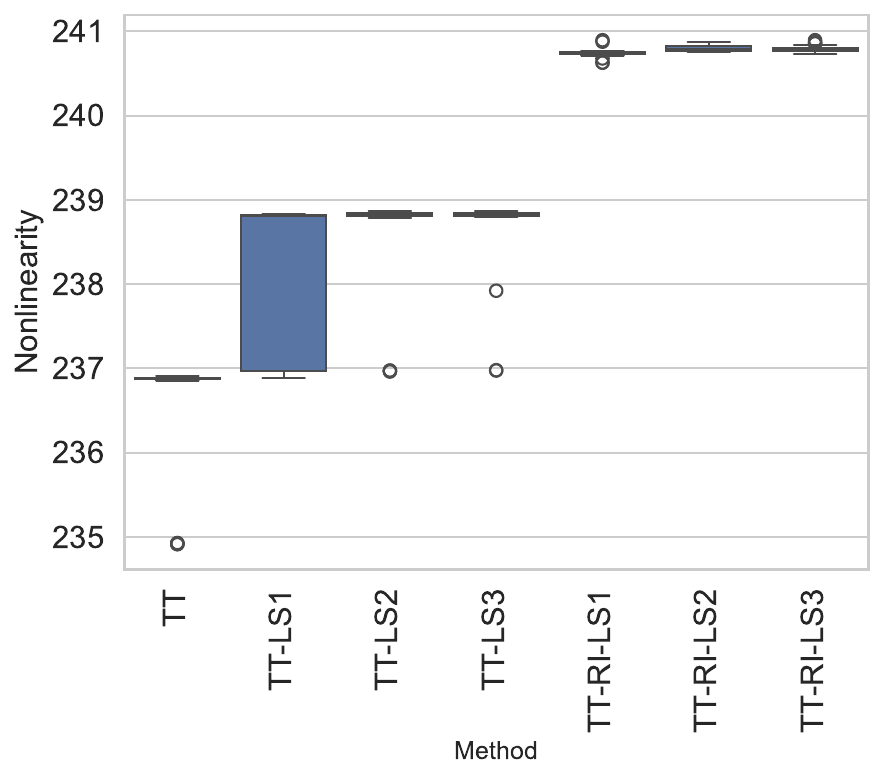}
    \caption{Boxplot representation of the results for the application of LS operators on the TT representation}
    \label{fig:boxls}
\end{figure}

\section{Conclusions and Future Work}
\label{sec:conclusions}

This paper systematically evaluates the evolution of Boolean functions with high nonlinearity and in odd sizes. The experiments with EAs and three solution encodings indicate GP to be the best option, regardless of the fact that GP works with general Boolean functions, while FP and TT consider only a subspace of rotation symmetric Boolean functions. 
Unfortunately, even such best results fall short of the best-known results reached with custom heuristics (except for the smallest Boolean function size). 
Next, we added several local search variants to our best EAs. Those modifications did not help GP but improved the TT results. Moreover, one combination of the TT encoding and local search operators even resulted in a nine variable Boolean function with nonlinearity 241, which is the best possible value within the rotation symmetric Boolean function class, and something previously not achieved with EAs.

Our results indicate several interesting future research directions. On the one hand, GP again shows to be the best general encoding, but with it, we cannot consider solutions belonging to a specific subclass of Boolean functions. It would be interesting to explore how to circumvent this problem. On the other hand, TT and FP are not very successful in the general case, but they benefit from constraining to subclasses and more choice of local search operators. It would be interesting to explore further diverse local search options one can use there.

\bibliographystyle{abbrv}
\bibliography{bibliography}

\end{document}